\documentclass[]{alaya}
\usepackage{makecell}
\usepackage{wrapfig}
\usepackage{tabularx}
\usepackage{textcomp}
\usepackage{stfloats}
\usepackage{url}
\usepackage{verbatim}
\usepackage{titlesec}
\usepackage{adjustbox}
\usepackage{multirow}
\usepackage{pifont}
\usepackage[sc]{mathpazo}
\usepackage{tikz}
\usepackage{comment}
\usepackage{amsmath,amssymb}
\usepackage{colortbl}
\usepackage{natbib}
\usepackage{color}
\usepackage{booktabs} 
\usepackage{hyperref}
\usepackage{graphicx}
\usepackage{multirow}
\RequirePackage{xspace}
\makeatletter
\DeclareRobustCommand\onedot{\futurelet\@let@token\@onedot}
\def\@onedot{\ifx\@let@token.\else.\null\fi\xspace}
\usepackage[most]{tcolorbox}
\usepackage{xcolor}
\usepackage{array}
\usepackage{tabularx}
\usepackage{siunitx} 
\usepackage{makecell}
\usepackage[table]{xcolor}
\definecolor{headerpurple}{HTML}{d8d2fc}
\definecolor{rowgray}{gray}{0.95}

\makeatother

\definecolor{adptorange}{RGB}{248, 205, 172}
\definecolor{cmpblue}{RGB}{189, 215, 238}
\definecolor{cmpblue}{RGB}{189, 215, 238}

\definecolor{our_red}{RGB}{232,157,160}
\definecolor{our_blue}{RGB}{136,206,230}
\definecolor{our_orange}{RGB}{246,200,168}
\definecolor{our_green}{RGB}{178,211,164}

\definecolor{attn_code0}{RGB}{247,215,200}
\definecolor{attn_code1}{RGB}{238,169,139}
\definecolor{mlp_code0}{RGB}{204,201,221}
\definecolor{mlp_code1}{RGB}{102,95,153}
\definecolor{mygray}{HTML}{f0f0f0}

\definecolor{token_blue}{RGB}{84, 120, 140}

\usepackage{pifont}
\usepackage{bbding}
\usepackage{fontawesome}
\usepackage{xspace}

\usepackage{float}

\newlength\savewidth

\newcolumntype{x}[1]{>{\centering\arraybackslash}p{#1pt}}
\newcolumntype{y}[1]{>{\raggedright\arraybackslash}p{#1pt}}
\newcolumntype{z}[1]{>{\raggedleft\arraybackslash}p{#1pt}}

\renewcommand{\paragraph}[1]{\vspace{1mm}\noindent\textbf{#1}}
\usepackage{colortbl}
\usepackage{xcolor}
\usepackage{wrapfig}

\renewcommand{\paragraph}[1]{\vspace{1.25mm}\noindent\textbf{#1}}

\usepackage{algorithm}
\usepackage{listings}

\definecolor{codeblue}{rgb}{0.25, 0.5, 0.5}
\definecolor{codekw}{rgb}{0.35, 0.35, 0.75}
\lstdefinestyle{Pytorch}{
    language = Python,
    backgroundcolor = \color{white},
    basicstyle = \fontsize{9pt}{8pt}\selectfont\ttfamily\bfseries,
    columns = fullflexible,
    aboveskip=1pt,
    belowskip=1pt,
    breaklines = true,
    captionpos = b,
    commentstyle = \color{codeblue},
    keywordstyle = \color{codekw},
}

\definecolor{green}{HTML}{009000}
\definecolor{red}{HTML}{ea4335}

\newcommand{\qstrip}[1]{\raisebox{-0.5\height}{\includegraphics[width=0.955\linewidth]{#1}}}
\newcommand{\qrowlab}[1]{\rotatebox[origin=c]{90}{\makecell{#1}}}
\newcommand{\qcolhdr}[1]{\makebox[0.191\linewidth]{#1}}

\newcommand{\qcolhdrfour}[1]{\makebox[0.2388\linewidth]{#1}}

\title{Generative World Renderer at the Speed of Play}
\author[1]{Guixu Lin}
\author[1]{Zheng-Hui Huang}
\author[1]{Siqi Yang}
\author[2]{Ming-Hsuan Yang}
\author[1]{Kaipeng Zhang}
\author[1,*]{Zhixiang Wang}

\affiliation[1]{Alaya Lab}
\affiliation[2]{University of California, Merced}
\demo{\url{https://alaya-renderer-flash.alayalab.ai}}
\correspondence{Zhixiang Wang <\email{zhixiang.wang@shanda.com}>}

\abstract{
Generative world renderer \textbf{AlayaRenderer} receives structured world states exported from physics engines and synthesizes RGB frames. Unlike models that generate frames from text/control-hints prompts, AlayaRenderer preserves scene structure without altering the underlying world dynamics. This demonstrates an alternative path toward interactive world modeling and user-controllable play. However, the original AlayaRenderer is too computationally expensive for real-time deployment.
This technical report introduces \textbf{AlayaRenderer-Flash}, a real-time-oriented generative forward world renderer that pushes AlayaRenderer from \textbf{0.56 FPS} to \textbf{31.54 FPS}, reaching the speed of play. AlayaRenderer-Flash reformulates the original renderer as a
few-step autoregressive streaming model and introduces lightweight distilled
codecs for efficient latent encoding and frame
reconstruction. It retains the teacher model's G-buffer and text-prompt interfaces while
enabling continuous rendering over input streams of unbounded length. We evaluate AlayaRenderer-Flash on G-buffer streams across content preservation, temporal consistency, cross-window stability, prompt controllability, and runtime efficiency. Our results show that AlayaRenderer-Flash substantially reduces inference
cost while preserving the core rendering capabilities of the teacher model.
By integrating AlayaRenderer-Flash with a physics engine, we build a fully
playable generative world running at 30 FPS.
}

\date{\today} 

\begin{document}
\maketitle

\begin{figure}[H]
  \vspace{-5pt}
  \centering
  \includegraphics[width=0.99\linewidth]{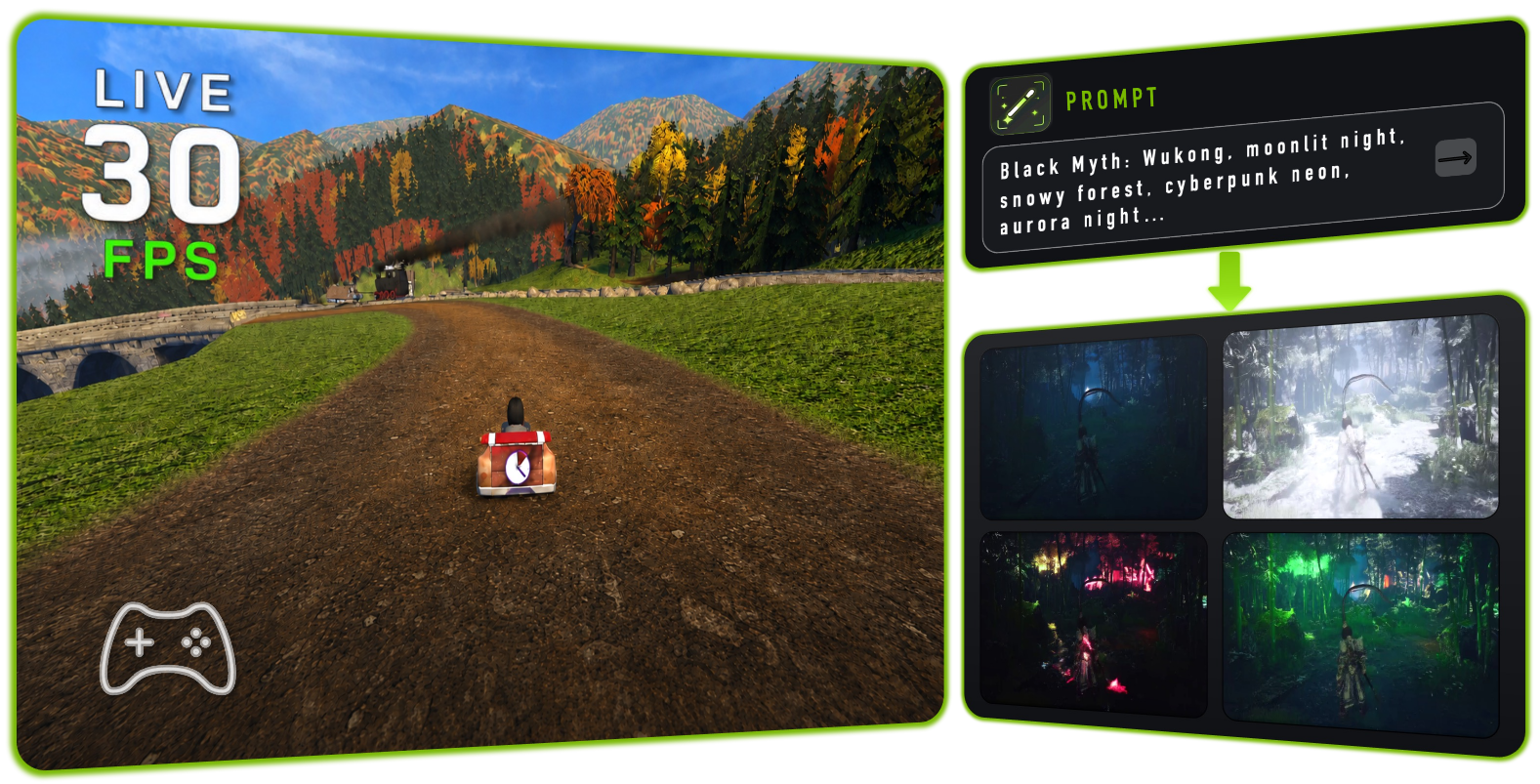}
  \vspace{3pt}
  \caption{AlayaRenderer-Flash restyling a live \emph{SuperTuxKart} session in our
interactive web demo. The game engine continuously exports synchronized
G-buffers to AlayaRenderer-Flash, which generates stylized RGB frames at
playback rate. Players can modify the visual style of the rendered game
through arbitrary text prompts while the underlying gameplay logic remains
unchanged.}
  \label{fig:teaser}
\end{figure}

\section{Introduction}

% \zhixiang{TODO: modify the story telling according to abstract...}
% In modern game engines, deferred rendering rasterizes the scene into physical
% and geometric buffers---such as depth, normals, albedo, roughness, and
% metallicity---before shading them into RGB frames. These G-buffers separate
% scene structure from final appearance
% \citep{saito1990comprehensible,nalbach2017deep,thies2019deferred}, and recent
% diffusion and flow renderers show that a generative model conditioned on them
% can take over the shading stage while respecting the given structure
% \citep{liang2025diffusion,beisswenger2025framediffuser,zhang2026renderflow}.

Generative models offer a promising path toward interactive visual worlds, but methods that synthesize frames solely from text prompts or control hints often struggle to preserve scene structure, material properties, and gameplay logic as the world evolves. A more principled alternative is to keep the physics engine responsible for simulating the world while delegating only the appearance generation to a generative renderer. Fortunately, modern game engines already expose a natural interface for this paradigm. In deferred rendering, scenes are first rasterized into physical and geometric buffers—including depth, normals, albedo, roughness, and metallicity—before being shaded into RGB frames \citep{saito1990comprehensible,nalbach2017deep,thies2019deferred}. These G-buffers encode the structured world state while leaving the final appearance unconstrained. Recent diffusion- and flow-based renderers have demonstrated that generative models conditioned on G-buffers can replace the conventional shading stage while faithfully respecting the underlying scene structure \citep{liang2025diffusion,beisswenger2025framediffuser,zhang2026renderflow}. Because geometry, physics, and gameplay logic remain under the control of the game engine, this engine-plus-generative-renderer paradigm preserves the evolving world by construction, providing a practical foundation for real-time, prompt-controllable gameplay.

AlayaRenderer, the forward renderer of the Generative World Renderer system \citep{huang2026generative}, represents the state of the art in this direction. Trained on a large-scale corpus captured from visually complex AAA games, it renders scenes far beyond the visual complexity handled by previous G-buffer renderers while supporting text-guided appearance control. Players can dynamically restyle the game—changing lighting, atmosphere, or artistic style—without modifying geometry, materials, or gameplay logic.
Despite its rendering quality, AlayaRenderer remains an offline renderer. Its 50-step denoising process dominates inference cost, while G-buffer encoding and VAE decoding introduce additional computational bottlenecks. Moreover, its bidirectional fixed-window formulation prevents autoregressive rollout over the unbounded G-buffer stream produced by a live game engine. As a result, prompt-controllable rendering cannot operate at interactive frame rates.

To bridge this gap, we introduce AlayaRenderer-Flash, a real-time deployment version of AlayaRenderer that preserves the original G-buffer and text-prompt interfaces while enabling interactive rendering through three complementary improvements. First, generation becomes autoregressive, allowing rendering history to be propagated across windows and enabling streaming over unbounded G-buffer sequences. Second, the original 50-step denoising schedule is distilled into a 4-step renderer, substantially reducing inference latency. Third, the computationally expensive G-buffer encoder and Wan VAE decoder~\citep{wan2025wan} are replaced with distilled lightweight encoder and decoder networks. Together, these improvements preserve the rendering quality and controllability of AlayaRenderer while reducing inference cost to the point where prompt-controllable generative rendering can run alongside live gameplay instead of as an offline post-processing pipeline.

\section{Method}
\label{sec:method}

\begin{figure}[!th]
\centering
\includegraphics[width=\linewidth]{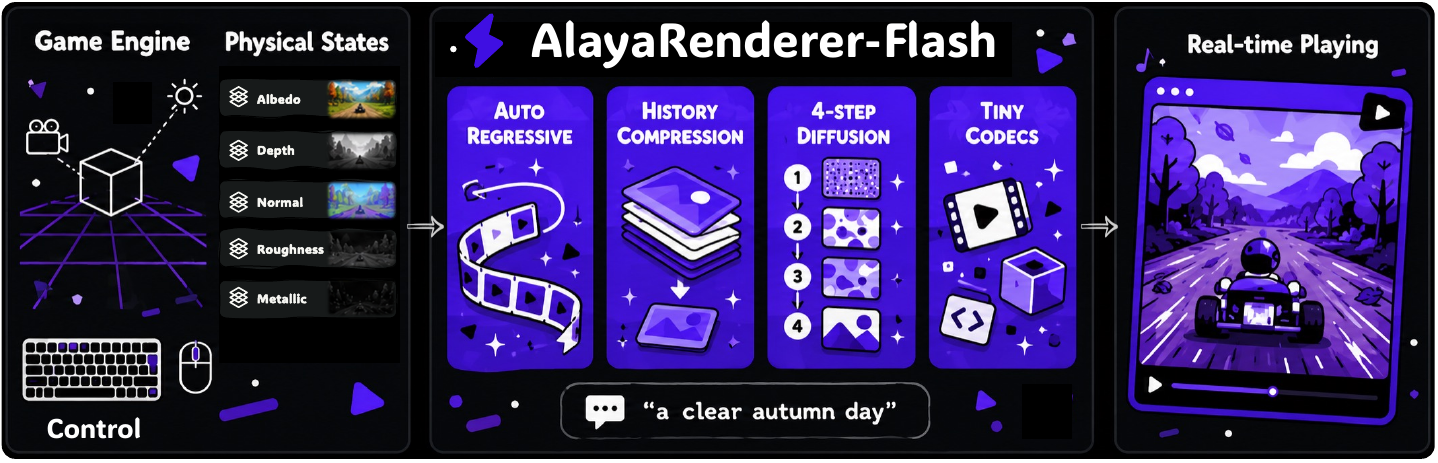}
\vspace{0.5pt}
\caption{System overview of AlayaRenderer-Flash. During gameplay, the player independently controls game interaction through standard game inputs and visual appearance through text prompts. The game engine continuously updates the world state and exports synchronized G-buffers, which, together with the text prompt, are processed by AlayaRenderer-Flash to generate stylized RGB frames in real time. The rendered frames are immediately displayed to the player, forming a closed-loop interactive rendering system.
% \zhixiang{Maybe draw this figure as a system @zhixiang}
}
\label{fig:framework}
\end{figure}

% AlayaRenderer-Flash preserves the original interface of
% AlayaRenderer while enabling real-time interactive deployment.
% As illustrated in Figure~\ref{fig:framework}, the player controls
% gameplay through standard game inputs and independently specifies the
% desired visual appearance using a text prompt. The game engine updates
% the world state according to player interactions and continuously exports
% synchronized G-buffers describing scene geometry and material
% properties. Together with the text prompt, these structured rendering
% buffers are fed into AlayaRenderer-Flash, which generates stylized RGB
% frames in real time. The rendered frames are immediately displayed to
% the player, forming a closed-loop interactive rendering system.

% Internally, AlayaRenderer-Flash consists of several key components.
% First, it reformulates the original bidirectional renderer as an
% autoregressive streaming model that rolls out over unbounded G-buffer
% streams while conditioning on its own generation history
% (Section~\ref{sec:method-causal}). Second, the original 50-step
% denoising process is distilled into a four-step student
% (Section~\ref{sec:method-distill}). Third, the original G-buffer encoder
% and VAE decoder are replaced with lightweight distilled modules for
% efficient latent encoding and frame reconstruction
% (Section~\ref{sec:method-decoder}). We describe each component in turn.
Figure~\ref{fig:framework} illustrates the complete deployment pipeline, in which a game engine continuously exports synchronized G-buffer streams that are processed by AlayaRenderer-Flash to generate stylized RGB frames in real time. Internally, AlayaRenderer-Flash combines autoregressive streaming with
hierarchical history compression, four-step diffusion, and lightweight
distilled codecs to enable playback-rate rendering. The following subsections
describe these designs in detail.

\subsection{G-Buffer-Conditioned Latent Rendering}
\label{sec:method-arch}
The base renderer, AlayaRenderer~\citep{huang2026generative}, is built upon
Wan~2.1~\citep{wan2025wan}, a latent video diffusion model that denoises RGB
video latents produced by the Wan causal 3D VAE while conditioning on text
through cross-attention. The game engine provides five synchronized
physical buffers---albedo, depth, metallic, normal, and
roughness---describing scene geometry and material properties rather than
final appearance.
Each G-buffer channel is treated as a video stream and encoded by the same
Wan causal 3D VAE as the RGB frames, yielding frame-aligned latent
representations. The resulting features are concatenated along the channel
dimension to form the G-buffer condition \(g_k\), which is concatenated
with the noisy RGB latent before the first 3D patch embedding:
\begin{equation}
    h_k=\operatorname{PatchEmbed}\!\left([x_k,g_k]\right),
    \label{eq:patchembed}
\end{equation}
where \(x_k\) denotes the noisy RGB latent of chunk \(k\), and \(g_k\) the
corresponding G-buffer condition.

This conditioning strategy requires only a minimal modification to the
pretrained backbone by widening the input projection of the first 3D patch
embedding to accommodate the additional condition channels. The remainder
of the diffusion transformer is left unchanged.

\subsection{Autoregressive Streaming}
\label{sec:method-causal}
 
AlayaRenderer performs bidirectional generation over a fixed-length window (21-latent-frame chunks)
and therefore cannot support unbounded live rendering. AlayaRenderer-Flash
instead operates autoregressively at the chunk level by partitioning the
latent video into fixed-length chunks.  In our experiments, each chunk
contains four latent frames, and previously generated chunks are retained as
history for subsequent generation.
We organize the retained history into a three-tier compression hierarchy,
with the compression level determined by its temporal distance from the
current chunk. Recent history is preserved at full fidelity, while more
distant history is progressively compressed into increasingly coarser
representations, following the general idea of multi-scale history
compression in recent
work~\citep{zhang2026frame,yuan2026helios}. The first generated latent frame
is additionally retained as a global appearance anchor by prepending it to
the highest-fidelity history tier, allowing every chunk to attend to the
stream's initial appearance. During denoising, all retained history is
inserted into self-attention together with the current chunk as clean latent
tokens, while the G-buffer latents remain the
per-chunk conditioning signals for geometry and material properties.

To preserve prompt controllability over rollouts of arbitrary length, we
further augment every self-attention layer with a dedicated text sink
consisting of persistent key--value entries projected from the style-prompt
embedding. Consequently, every chunk continuously attends to the prompt
through self-attention in addition to the standard cross-attention
conditioning.

The full encoding--decoding pipeline is likewise implemented causally: the VAE encoder carries its temporal feature cache across chunks so the G-buffer stream is encoded as one continuous sequence, and the decoder's temporal state persists across chunk calls, so that per-chunk streaming decoding is numerically consistent with a single-pass decode of the entire latent stream.

\subsection{Few-Step Distillation}
\label{sec:method-distill}

Reducing the number of denoising steps is one of the most effective ways
to accelerate diffusion inference. We therefore target a four-step
denoising schedule for deployment and use it throughout all experiments.
However, directly compressing the original 50-step denoising process into
four steps causes substantial quality degradation, since each denoising
step must approximate a much longer trajectory than the teacher was
trained to model.

To overcome this challenge, we adopt a three-stage few-step distillation
pipeline consisting of guidance distillation, progressive step reduction,
and Mean Flow Distillation (MFD)~\citep{zhao2026mean}. We find that the
final distribution-refinement stage is highly sensitive to the student's
initialization: directly applying MFD to an ODE-regression-initialized
four-step student leads to unstable optimization. The first two stages
therefore provide a stable initialization by progressively adapting the
student to the target four-step budget before the final refinement stage. 

\paragraph{Guidance Distillation.}
The autoregressive renderer introduced in
Section~\ref{sec:method-causal} still relies on classifier-free guidance
(CFG), requiring two network evaluations at every denoising step. We
therefore distill the teacher's CFG-combined velocity field at the target
guidance scale into the student~\citep{meng2023distillation}. During this
stage, the student retains the original 50-step denoising schedule while
learning to reproduce the teacher's guided prediction with a single
network evaluation. This converts explicit CFG into behavior encoded in
the student weights, enabling all subsequent distillation stages to use
the same single-pass formulation.

\paragraph{Progressive Step Reduction.}
To provide a stable initialization for the final distribution-refinement
stage, we gradually reduce the denoising budget following Progressive
Distillation~\citep{salimans2022progressive}. Directly training the student
under the target four-step budget leads to unstable optimization.
Therefore, we sequentially train the guidance-distilled student using
32-, 16-, 8-, and finally 4-step schedules, allowing it to progressively
adapt to increasingly larger denoising intervals and providing a stable
initialization for the final refinement stage.

\paragraph{Mean Flow Distillation under self-rollout.}
After progressive distillation, the four-step student is further refined
under the self-rollout training regime introduced in Self
Forcing~\citep{huang2026self}. During this stage, the student conditions
exclusively on its own previously generated chunks rather than
ground-truth RGB history, matching the deployment setting and reducing
the train--test discrepancy between teacher-forced training and
autoregressive inference.

For the final distribution-refinement stage, we experimented with
DMD-style adversarial distribution
matching~\citep{yin2024improved}, but found it prone to unstable
optimization and noticeable color artifacts in our setting. We therefore
adopt MFD~\citep{zhao2026mean}, whose flow-based
matching objective provides substantially more stable optimization for
four-step refinement while maintaining comparable visual fidelity.

Finally, aggressive step reduction inevitably suppresses high-frequency
details. To compensate for this loss, we attach lightweight GAN heads to
intermediate transformer features and the final latent prediction
(MFD+GAN). The adversarial loss is assigned a small weight to recover
local textures without interfering with the MFD objective or destabilizing
training.

\subsection{Distilled Tiny Codecs}
\label{sec:method-decoder}

After reducing denoising to only a few diffusion steps, the remaining
inference latency is dominated by the VAE modules at both ends of the
pipeline. Specifically, encoding requires five independent Wan VAE passes,
one for each G-buffer channel, while decoding relies on the streaming Wan
VAE decoder. To further reduce latency, we distill both components into
lightweight replacements: a shared tiny G-buffer encoder and a tiny temporal
decoder. 
% Throughout the paper, \emph{4-step distilled} refers to the
% intermediate configuration that still uses the original Wan VAE encoder and
% decoder, whereas \emph{AlayaRenderer-Flash} denotes the final deployment
% configuration with both distilled modules.

\paragraph{Tiny decoder.}
The tiny decoder adopts the architecture of TAEHV~\citep{taehv2025} and is
initialized from its publicly released Wan~2.1~\citep{wan2025wan} pretrained weights. However,
these pretrained weights do not reconstruct our game-domain content
sufficiently well. We therefore further distill the tiny decoder from the
frozen Wan VAE decoder. During training, both decoders receive the same
latent representations produced by the diffusion transformer, and the tiny
decoder is optimized to reproduce the outputs of the Wan VAE decoder using a
pixel reconstruction loss together with a small perceptual
loss~\citep{zhang2018unreasonable}.  

\paragraph{Tiny G-buffer encoder.}
The input encoder is distilled in the same manner. Instead of performing
five separate Wan VAE encoding passes, one for each G-buffer channel, we
train a shared tiny G-buffer encoder that predicts the G-buffer condition
\(g_k\) in Equation~\ref{eq:patchembed} with a single forward pass. The
encoder is first distilled to match the latent representations produced by
the frozen Wan VAE encoder, and is then jointly fine-tuned together with the
entire renderer.

% \paragraph{Tiny G-buffer encoder.}
% The input side uses the same distillation recipe: the five per-buffer Wan
% VAE encoding passes are distilled into a compact shared tiny G-buffer encoder
% that produces the same G-buffer condition \(g_k\) as in
% Equation~\ref{eq:patchembed} in a single forward pass, trained first to
% match the frozen VAE latents and then fine-tuned jointly with the renderer.

\subsection{Real-Time Deployment}
\label{sec:method-deploy}

The techniques introduced above are integrated into a unified streaming
renderer for real-time deployment. During inference, incoming G-buffer
streams are continuously encoded by the distilled tiny encoder, rendered
by the four-step diffusion model, and decoded into RGB frames by the
stateful tiny decoder. All components operate autoregressively while
maintaining persistent temporal states across chunk boundaries, enabling
continuous rendering over unbounded input streams without restarting the
rendering pipeline.

The complete inference pipeline is executed as a single optimized
streaming system, minimizing runtime overhead while preserving temporal
state throughout continuous execution. Together, the autoregressive
streaming formulation, four-step diffusion model, and lightweight codecs
enable playback-rate rendering suitable for interactive applications.

\section{Experiments}
\label{sec:setup}

\paragraph{Data \& Experimental Setup.}
AlayaRenderer-Flash is trained and evaluated on an engine-captured
\textit{Black Myth: Wukong} dataset collected at a resolution of
\(1280\times720\) and 30 FPS. The dataset consists of 1,352 training clips
and 131 test clips. Each frame contains an RGB reference together with five
aligned G-buffer channels: albedo, depth, normal, roughness, and metallic.
The data are collected using the same capture pipeline as the Generative
World Renderer dataset~\citep{huang2026generative}. Unless otherwise
specified, all quantitative results are reported on the 131 test clips,
each containing 150 frames (5 seconds), at a model input resolution of
\(832\times448\). All stages of training are performed on eight NVIDIA H200 GPUs.

% \paragraph{Compared configurations.}
% Table~\ref{tab:main} compares four configurations within the
% AlayaRenderer family. {AlayaRenderer} is the original
% bidirectional renderer using a 50-step denoising schedule.
%  {AlayaRenderer-AR} replaces the bidirectional formulation with the
% autoregressive streaming introduced in
% Section~\ref{sec:method-causal} while retaining the original 50-step
% sampling schedule.
% {AlayaRenderer-AR-distilled} further replaces the 50-step sampler
% with the proposed 4-step distilled model. Finally,
% {AlayaRenderer-Flash} combines the distilled renderer with the
% lightweight encoder and decoder introduced in
% Section~\ref{sec:method-decoder}, representing the complete proposed
% method.

% \paragraph{Protocol.}
% All compared configurations use identical G-buffer inputs, text prompts,
% resolution, and preprocessing. Autoregressive models are evaluated by
% rolling out their own generated history without injecting ground-truth RGB
% frames, matching the intended deployment setting where prediction errors can
% accumulate over time. Consequently, AlayaRenderer serves as a high-quality
% reference rather than a real-time baseline. For external comparisons, only
% methods with reproducible implementations under the same G-buffer protocol
% are included quantitatively; other published systems are discussed
% qualitatively.
\paragraph{Metrics.}
We evaluate rendering quality and deployment performance using the
following complementary metrics.

\begin{itemize}

\item \textbf{Content preservation.}
We report CLIP image-image similarity
(\(S_{\mathrm{CLIP-I}}\)) using the OpenCLIP ViT-L/14 image encoder
\citep{radford2021learning}. This metric measures semantic consistency
between generated frames and their paired reference frames while remaining
relatively insensitive to changes in lighting, color, and artistic style.

% \item \textbf{Overall video quality.}
% We report Fréchet Video Distance (FVD)
% \citep{unterthiner2019fvd}, which measures the distributional similarity
% between generated and reference video clips in a pretrained video feature
% space. Lower FVD indicates higher overall video quality and better
% spatiotemporal fidelity.

\item \textbf{Temporal stability.}
% Temporal consistency is evaluated using warped temporal LPIPS
% (\(\mathrm{tLPIPS}_{\mathrm{warp}}\))
% \citep{zhang2018unreasonable}, where lower values indicate reduced
% perceptual flicker between consecutive generated frames.
Temporal consistency is evaluated using warped temporal LPIPS
(\(\mathrm{tLPIPS}_{\mathrm{warp}}\))
\citep{zhang2018unreasonable}. 
For each adjacent generated-frame pair, the previous frame is first aligned to the current frame using dense optical flow, and LPIPS is then computed between the aligned previous frame and the current frame.
Lower values indicate less perceptual flicker.

\item \textbf{Cross-window consistency.}
To evaluate long-horizon streaming, we report Boundary MSE and Boundary
SSIM across adjacent evaluation windows following the protocol of
\citep{yang2026horizonrelight}. Lower Boundary MSE and higher Boundary
SSIM indicate smoother transitions between independently processed
streaming windows.

\item \textbf{Prompt controllability.}
We quantify prompt controllability with a contrastive CLIP margin. For each
generated frame sampled at a fixed stride, we compare its CLIP similarity to
the prompt used for generation against its similarity to a contrast prompt from
another test clip:
\[
M_{\mathrm{CLIP}}
=
\operatorname{CLIP}(I,y^{+})
-
\operatorname{CLIP}(I,y^{-}),
\]
where \(y^{+}\) is the conditioning prompt and \(y^{-}\) is the contrast
prompt. The score is averaged over frames and clips. Higher values indicate
stronger text control over rendered appearance, such as lighting, color,
weather, and atmosphere. For prompt-switch evaluation, the previous prompt is
used as \(y^{-}\). Methods without text conditioning are marked as not
applicable.

\item \textbf{Deployment performance.}
We report throughput (FPS) and peak GPU memory on an NVIDIA H200 GPU
under identical runtime settings.

\end{itemize}

\subsection{Progressive Design Analysis}
\label{sec:main-comparison}

\begin{table}[t]
\centering
\caption{
Comparison within the AlayaRenderer family. Bold
marks the best value per column.
}
\label{tab:main}
\resizebox{\linewidth}{!}{
\begin{tabular}{lccccccccc}
\toprule
Method
& Steps
& \(S_{\text{CLIP-I}}\) $\uparrow$
& \makecell{Boundary\\MSE $\downarrow$}
& \makecell{Boundary\\SSIM $\uparrow$}
& \(\mathrm{tLPIPS}_{\mathrm{warp}}\) $\downarrow$
& \(M_{\mathrm{CLIP}}\) $\uparrow$
& FPS $\uparrow$
& VRAM (GB) $\downarrow$ \\
\midrule
AlayaRenderer
& 50 & 0.836 & 0.0500 & 0.308 & \textbf{0.124} & 0.039
&  0.56 & 30.1 \\
AlayaRenderer-AR
& 50 & 0.846 & 0.0433 & 0.358 & 0.197 & 0.030
& 1.53 & 22.6 \\
 AlayaRenderer-AR-distilled
& 4 & 0.843 & 0.0418 & 0.371 & 0.158 & \textbf{0.045}
&  6.30 & 22.6 \\
\textbf{AlayaRenderer-Flash}
& 4 & \textbf{0.847} & \textbf{0.0406} & \textbf{0.430}
& 0.155 & 0.043
& \textbf{31.54} & \textbf{16.2} \\
\bottomrule
\end{tabular}
}
\end{table}

Table~\ref{tab:main} presents a controlled progression from the original
AlayaRenderer to the final AlayaRenderer-Flash through four configurations:
(1) the original AlayaRenderer,
(2) autoregressive streaming with the original 50-step denoising schedule,
(3) 4-step distilled rendering, and
(4) lightweight deployment with the distilled encoder and decoder.
All configurations are evaluated under the same 5-second protocol using
identical G-buffer inputs, text prompts, and evaluation settings, allowing
the contribution of each stage to be isolated.

% \paragraph{Autoregressive reformulation.}
% AlayaRenderer-AR isolates the effect of replacing the original
% bidirectional formulation with the proposed autoregressive streaming
% formulation while retaining the original 50-step denoising schedule.
% Compared with AlayaRenderer, it achieves higher content similarity
% (\(S_{\mathrm{CLIP-I}}\): 0.846 vs.\ 0.836) together with improved
% boundary consistency (Boundary MSE: 0.0433 vs.\ 0.0500; Boundary SSIM:
% 0.358 vs.\ 0.308). Temporal stability decreases
% (\(\mathrm{tLPIPS}_{\mathrm{warp}}\): 0.197 vs.\ 0.124), reflecting the
% more challenging setting of autoregressive streaming. Throughput also
% improves from 0.56 FPS to 1.53 FPS, but remains well below the target
% deployment rate, indicating that autoregressive streaming alone is
% insufficient for real-time rendering.

\paragraph{Autoregressive reformulation.}
AlayaRenderer-AR isolates the effect of the proposed autoregressive
streaming formulation while retaining the original 50-step denoising
schedule. Compared with AlayaRenderer, it achieves higher content
similarity (\(S_{\mathrm{CLIP-I}}\): 0.846 vs.\ 0.836) and improved
boundary consistency (Boundary MSE: 0.0433 vs.\ 0.0500; Boundary SSIM:
0.358 vs.\ 0.308). Temporal stability decreases
(\(\mathrm{tLPIPS}_{\mathrm{warp}}\): 0.197 vs.\ 0.124), reflecting the
more challenging autoregressive setting. Throughput increases from
0.56 FPS to 1.53 FPS, but remains far below the target playback rate,
indicating that autoregressive streaming alone is insufficient for
real-time deployment.

\paragraph{4-step distillation.}
Replacing the original 50-step denoising process with the proposed
4-step distilled model substantially improves computational efficiency while
largely preserving rendering quality. Relative to the 50-step
autoregressive renderer, throughput increases from 1.53 FPS to 6.30 FPS
without increasing peak memory usage. Meanwhile, content similarity,
boundary consistency, and prompt controllability remain comparable to, or
slightly better than, the 50-step model. Temporal stability also improves
considerably, with warped temporal LPIPS decreasing from 0.197 to 0.158.
These results demonstrate that 4-step distillation substantially improves
computational efficiency while maintaining high rendering quality.

\paragraph{Lightweight deployment.} 
AlayaRenderer-Flash further replaces the original G-buffer encoder and Wan
VAE decoder with their distilled lightweight counterparts while keeping the
same four-step diffusion transformer. Consequently, this comparison
isolates the contribution of the lightweight codec modules to deployment
efficiency. The lightweight implementation increases throughput from
6.30 FPS to 31.54 FPS while reducing peak GPU memory from 22.6 GB to
16.2 GB. Meanwhile, rendering quality is maintained or slightly improved
across content similarity, and boundary
consistency. We attribute these gains to the lightweight encoder and
decoder being distilled and fine-tuned on the target game domain. These results indicate that the lightweight
codecs remove the remaining deployment bottleneck with negligible impact
on rendering quality.

\subsection{Capability and Performance Comparison}
\label{sec:external-comparison}

\begin{table}[t]
\centering
\caption{
Capability comparison with existing G-buffer rendering methods.
}
\label{tab:capability}
\resizebox{1\linewidth}{!}{
\begin{tabular}{lccccc}
\toprule
Method
& G-buffer
& Autoregressive
& Prompt switch
& Few-step
& Length \\
\midrule
RGB\(\leftrightarrow\)X (per-frame)
& yes & no & yes & no & per-frame \\
FrameDiffuser (per-frame AR)
& yes & yes & no & no & unbounded \\
\textbf{AlayaRenderer-Flash}
& yes & yes & yes & yes & unbounded  \\
\bottomrule
\end{tabular}
}
\end{table}

\begin{table}[t]
\centering
\caption{
Quantitative comparison with external G-buffer rendering
baselines under the fixed 5-second evaluation protocol.
Bold indicates the best result.
}
\label{tab:external}
\resizebox{\linewidth}{!}{
\begin{tabular}{llccccccc}
\toprule
Method
& Window
& \(S_{\text{CLIP-I}}\) $\uparrow$
& FVD $\downarrow$
& \(\mathrm{tLPIPS}_{\mathrm{warp}}\) $\downarrow$
& \(M_{\mathrm{CLIP}}\) $\uparrow$
& FPS $\uparrow$
& VRAM (GB)$\downarrow$ \\
\midrule
RGB\(\leftrightarrow\)X
& 5s & 0.820 & 1031.3 & 0.305 & 0.027 & 1.30 & \textbf{3.3} \\
FrameDiffuser 
& 5s & 0.844 & 650.6 & 0.440 & \texttt{n/a}  & 0.31 & 3.5 \\
\textbf{AlayaRenderer-Flash}
& 5s & \textbf{0.847} & \textbf{384.1}
& \textbf{0.155} & \textbf{0.043} & \textbf{31.54} & 16.2 \\
\bottomrule
\end{tabular}
}
\end{table}

For a fair comparison, we retrain all external baselines on our
training dataset using the same G-buffer benchmark whenever their
implementations are publicly available and reproducible.
Table~\ref{tab:capability} summarizes the capabilities of different
methods, while Table~\ref{tab:external} reports quantitative comparisons
under the same 5-second evaluation protocol.
RGB$\leftrightarrow$X~\citep{zeng2024rgbx} performs independent per-frame
rendering without temporal modeling. Consequently, it achieves the lowest
content similarity and the highest FVD~\citep{unterthiner2018towards}  among the evaluated methods,
indicating limited spatiotemporal coherence despite supporting text-guided appearance control.
FrameDiffuser~\citep{beisswenger2025framediffuser} introduces
autoregressive temporal modeling and improves content similarity over
RGB$\leftrightarrow$X. However, it does not support prompt-controlled
rendering and exhibits substantially weaker temporal stability
(\(\mathrm{tLPIPS}_{\mathrm{warp}}=0.440\)). Moreover, its inference
speed is limited to 0.31 FPS, making real-time streaming deployment
impractical.

We additionally retrain 
DiffusionRenderer~\citep{liang2025diffusion} on our training dataset.
Since DiffusionRenderer adopts a bidirectional formulation rather than
autoregressive streaming, each 150-frame evaluation sequence is divided
into multiple independent 25-frame windows, which are rendered offline
and evaluated separately. This protocol allows DiffusionRenderer to
exploit future context within each window, providing a favorable
evaluation setting for bidirectional methods. Under this protocol, it
achieves a higher \(S_{\mathrm{CLIP-I}}\) of 0.870 and a lower FVD of
335.5 than AlayaRenderer-Flash. However, its temporal stability remains
comparable (\(\mathrm{tLPIPS}_{\mathrm{warp}}=0.156\)), while inference
runs at only 1.10 FPS, preventing real-time streaming deployment.

Overall, AlayaRenderer-Flash achieves the best trade-off between
rendering quality and deployment efficiency. Although bidirectional
methods can obtain stronger offline rendering quality under favorable
evaluation protocols, AlayaRenderer-Flash is the only method that
simultaneously supports causal streaming, prompt switching, few-step
inference, and real-time interactive rendering while sustaining
31.54 FPS.

\subsection{Qualitative Results}
\label{sec:qualitative}

\begin{figure}[tb!]
\centering
\scriptsize
\newcommand{\qframehdr}{\qcolhdr{frame 0}\qcolhdr{frame 37}\qcolhdr{frame 74}\qcolhdr{frame 112}\qcolhdr{frame 140}}
\newcommand{\qframehdrshort}{\qcolhdr{frame 89}\qcolhdr{frame 94}\qcolhdr{frame 99}\qcolhdr{frame 104}\qcolhdr{frame 109}}
\setlength{\tabcolsep}{0pt}
\begin{tabular}{@{}c@{\hspace{2pt}}c@{}}
 & \qframehdr \\[1pt]
\qrowlab{Reference} & \qstrip{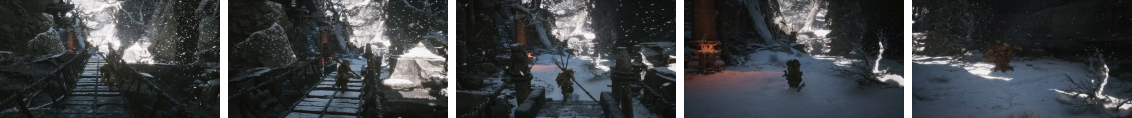} \\[2pt]
\qrowlab{RGB$\leftrightarrow$X} & \qstrip{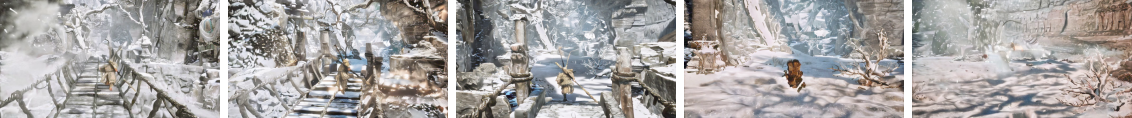} \\[2pt]
\qrowlab{FrameDiffuser} & \qstrip{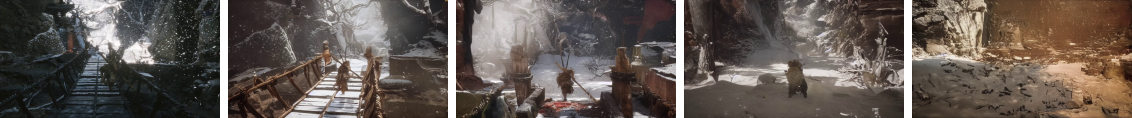} \\[2pt]
\qrowlab{Ours} & \qstrip{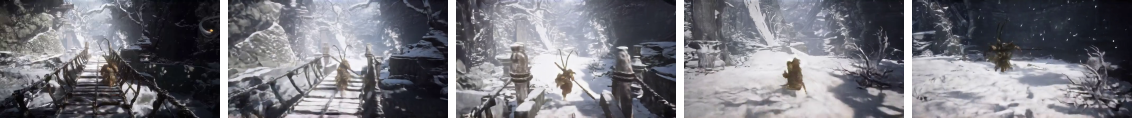} \\[2pt]
\\
 & \qframehdrshort \\[1pt]
\qrowlab{Reference} & \qstrip{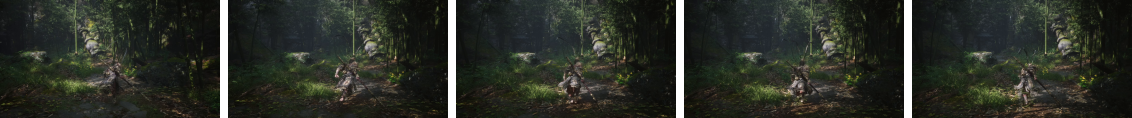} \\[2pt]
\qrowlab{RGB$\leftrightarrow$X} & \qstrip{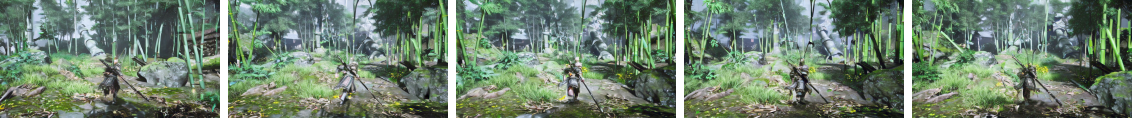} 
\\[2pt]
\qrowlab{FrameDiffuser} & \qstrip{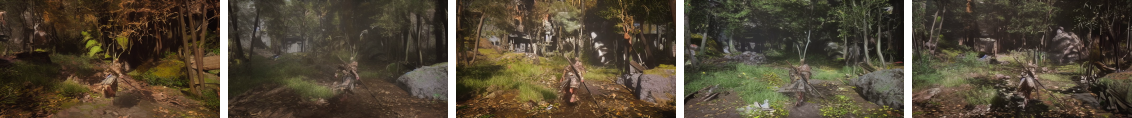} \\[2pt]
\qrowlab{Ours} & \qstrip{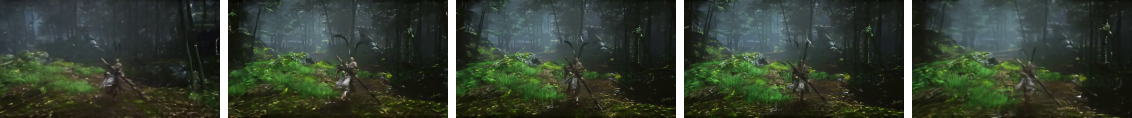} \\
\end{tabular}
\vspace{5pt}
\caption{Qualitative comparison with external baselines on 5-second clips. Reference denotes the original RGB frames rendered by the game engine during data collection. Since prompt-conditioned rendering admits multiple valid outputs, these frames serve as visual references rather than ground-truth targets.
}
\label{fig:qualitative-sequence}
\end{figure}

Figure~\ref{fig:qualitative-sequence} complements the quantitative
evaluation with representative examples from two challenging
sequences. The top block samples a snow scene across the full 5-second
rollout to show long-horizon behavior; the bottom block samples a bamboo
forest every five frames to expose short-term stability.

The qualitative observations are consistent with the quantitative results
reported in Table~\ref{tab:external}. RGB$\leftrightarrow$X~\citep{zeng2024rgbx} performs
independent per-frame rendering without temporal modeling and therefore
exhibits inconsistent illumination and visible temporal flicker throughout
the sequence. FrameDiffuser~\citep{beisswenger2025framediffuser} preserves more scene structure than
RGB$\leftrightarrow$X~\citep{zeng2024rgbx} but accumulates appearance drift during long
autoregressive rollouts, leading to noticeable changes in lighting, color
tone, and geometry over time. In contrast, AlayaRenderer-Flash maintains
stable illumination, scene geometry, and camera motion throughout the
entire rollout while remaining closely aligned with the underlying
G-buffer stream. These qualitative results demonstrate its ability to
preserve long-horizon temporal consistency while maintaining high visual
fidelity, consistent with the quantitative evaluation in
Table~\ref{tab:external}.

\subsection{Prompt Switching on Long Rollouts}
\label{sec:prompt-switch}

\begin{figure}[t]
\centering
\scriptsize
\setlength{\tabcolsep}{0pt}
\begin{tabular}{@{}c@{\hspace{2pt}}c@{}}
 & \qcolhdrfour{\textbf{cyberpunk}}\qcolhdrfour{\textbf{volcanic}}\qcolhdrfour{\textbf{sandstorm}}\qcolhdrfour{\textbf{arctic}} \\[1pt]
\qrowlab{Reference} & \qstrip{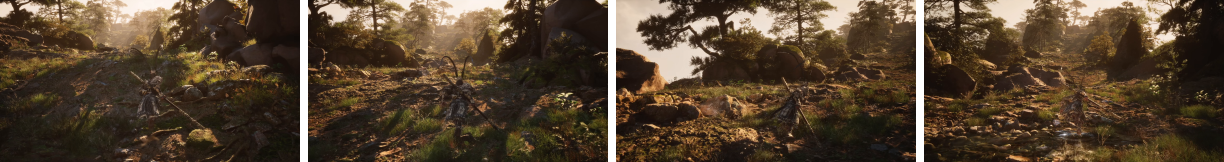} \\[2pt]
\qrowlab{Ours} & \qstrip{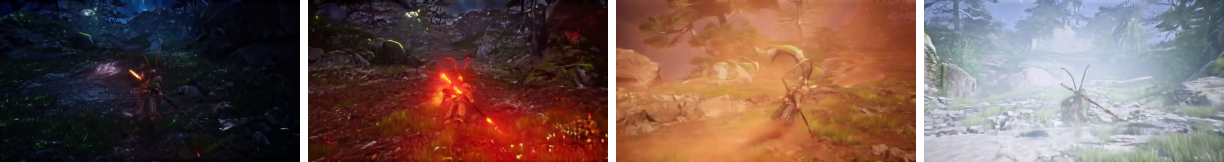} \\[8pt]
\\
 & \qcolhdrfour{\textbf{vaporwave}}\qcolhdrfour{\textbf{divine gold}}\qcolhdrfour{\textbf{deep-sea}}\qcolhdrfour{\textbf{infrared}} \\[1pt]
\qrowlab{Reference} & \qstrip{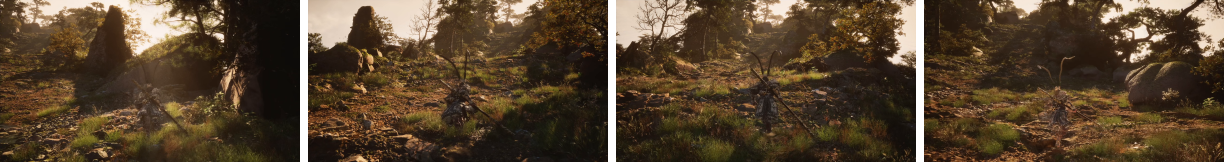} \\[2pt]
\qrowlab{Ours} & \qstrip{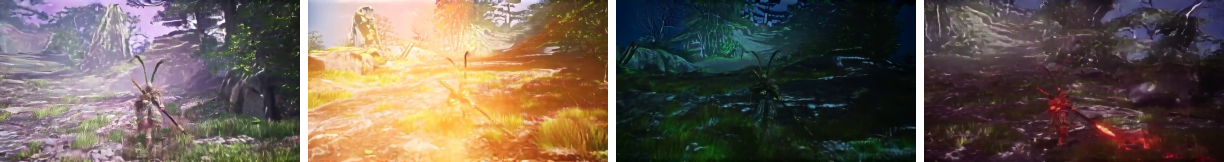} \\
\end{tabular}
\vspace{5pt}
\caption{
Prompt-controlled visual style transitions during a single long streaming
rollout. Reference denotes the original RGB frames rendered by the game engine during data collection and is shown for visual comparison.
}
\label{fig:prompt-switch-8}
\end{figure}
Prompt-controllable rendering is a key capability of
AlayaRenderer-Flash. Unlike offline renderers that require restarting the
generation process after each prompt change, our method supports seamless
prompt updates during continuous streaming inference.
We further evaluate this capability on longer test sequences, each
consisting of 637 frames rendered in a single streaming pass. During
generation, the prompt is switched every five chunks, cycling through
eight prompts spanning diverse artistic styles, lighting conditions,
weather effects, and color palettes.
Figure~\ref{fig:prompt-switch-8} shows a representative rollout.
AlayaRenderer-Flash consistently follows each prompt transition while
preserving scene geometry, character motion, and camera trajectory
throughout the sequence. Transitions remain smooth without noticeable
ghosting or boundary artifacts, demonstrating continuous
prompt-controlled rendering over long streaming rollouts.

\subsection{End-to-end Interactive Deployment}

To validate practical deployment, we adapt AlayaRenderer-Flash to the open-source game SuperTuxKart~\citep{supertuxkart}. We first collect a target-domain dataset containing synchronized RGB frames and the same five G-buffer modalities used by the renderer, and fine-tune the model on this dataset. We then integrate the fine-tuned renderer into the live SuperTuxKart engine to form a complete real-time interactive rendering system. During gameplay, players control vehicle motion and camera viewpoints through standard game inputs while independently specifying the desired visual appearance using predefined style presets or arbitrary text prompts. The game engine continuously updates the world state and exports synchronized G-buffer streams, which are processed by AlayaRenderer-Flash to synthesize stylized RGB frames while preserving the engine-side gameplay logic, physics simulation, scene geometry, and camera motion. Prompts can be updated online without interrupting gameplay, enabling continuous visual-style changes during interaction.

The complete rendering pipeline is deployed on a single NVIDIA H200 GPU. Under this deployment setting, AlayaRenderer-Flash—including G-buffer encoding, four-step diffusion, and RGB decoding—runs at 31.54 FPS. Including engine-side G-buffer readback, data transfer, and display synchronization, the complete interactive system consistently sustains the target playback rate of 30 FPS during live gameplay. These results demonstrate that AlayaRenderer-Flash can be deployed as a practical interactive renderer, supporting continuous prompt-controlled world rendering during live gameplay.
\section{Conclusion}

We presented AlayaRenderer-Flash, a real-time deployment framework that
transforms AlayaRenderer from an offline diffusion renderer into a
streaming generative renderer for interactive applications. By combining
autoregressive streaming, progressive four-step distillation, and
lightweight codecs, AlayaRenderer-Flash enables real-time rendering while
preserving the rendering quality, prompt controllability, temporal
consistency, and structural fidelity of the original model.

Extensive experiments demonstrate that AlayaRenderer-Flash achieves a
strong balance between rendering quality and deployment efficiency,
supporting continuous prompt-controlled rendering over arbitrarily long
streaming sequences at real-time speed. By integrating
AlayaRenderer-Flash into a game engine, we further demonstrate a fully
playable real-time game, validating the practical feasibility of
streaming generative world rendering.

We hope this work serves as a practical step toward real-time
generative world models and AI-native interactive environments.

\bibliographystyle{assets/plainnat}
\bibliography{references}

\end{document}